\documentclass[11pt]{article}
\usepackage{nodalida2025}
\usepackage{times}
\usepackage[draft]{hyperref}
\usepackage{latexsym}
\usepackage{amsmath}
\usepackage{tabularx}
\usepackage{listings}
\aclfinalcopy 
\usepackage{placeins}

\title{Efficient Scientific Full Text Classification: The Case of EICAT Impact Assessments}

\author{Marc Brinner \\
  Computational Linguistics \\
  Bielefeld University, Germany \\
  {\tt marc.brinner@uni-bielefeld.de} \\\And
  Sina Zarrieß \\
  Computational Linguistics \\
  Bielefeld University, Germany \\
  {\tt sina.zarriess@uni-bielefeld.de} \\}
  \author{Marc Brinner \and Sina Zarrieß \\
  Computational Linguistics, Department of Linguistics\\
  Bielefeld University, Germany\\
  \texttt{\{marc.brinner,sina.zarriess\}@uni-bielefeld.de}}

\date{}

\begin{document}
\maketitle
\begin{abstract}

This study explores strategies for efficiently classifying scientific full texts using both small, BERT-based models and local large language models like Llama-3.1 8B. We focus on developing methods for selecting subsets of input sentences to reduce input size while simultaneously enhancing classification performance. To this end, we compile a novel dataset consisting of full-text scientific papers from the field of invasion biology, specifically addressing the impacts of invasive species. These papers are aligned with publicly available impact assessments created by researchers for the International Union for Conservation of Nature (IUCN). Through extensive experimentation, we demonstrate that various sources like human evidence annotations, LLM-generated annotations or explainability scores can be used to train sentence selection models that improve the performance of both encoder- and decoder-based language models while optimizing efficiency through the reduction in input length, leading to improved results even if compared to models like ModernBERT that are able to handle the complete text as input. Additionally, we find that repeated sampling of shorter inputs proves to be a very effective strategy that, at a slightly increased cost, can further improve classification performance.

\end{abstract}

\section{Introduction}

The exponential growth of research publications across various domains \cite{bornmann_growth_2021} has created an increasing need for automated methods to process scientific texts efficiently. To address this, numerous approaches have been developed to optimize general research workflows, such as literature search \cite{singh-etal-2023-scirepeval} and summarization \cite{singha-roy-mercer-2024-enhancing}. For more specialized tasks, such as extracting specific information from full texts, proprietary large language models (LLMs) offer potential solutions \cite{dagdelen2024structured}. However, these models are not locally deployable, making them expensive to use when processing large datasets.

Recently, open-source large language models have emerged as strong competitors to proprietary systems, offering comparable performance \cite{deepseekai2024deepseekv3technicalreport}. Nevertheless, a wider adoption from researchers outside the machine learning research community is unlikely within the next years, primarily due to their significant hardware requirements. Furthermore, both proprietary and open-source LLMs of this scale are highly energy-intensive, raising concerns about their sustainability. This highlights the importance of exploring smaller, more efficient models that can deliver similar performance while minimizing resource consumption, or of exploring other strategies to reduce the computational cost of using these models to solve specific tasks.

To address these challenges, we investigate the potential of more efficient BERT-based models alongside slightly more resource-intensive local large language models (LLMs) for classification of scientific full texts. As part of this effort, we introduce the EICAT dataset, which consists of scientific full-text papers focused on specific invasive species and their impact on the native ecosystem, as well as labels specifying the impact category of that species with corresponding evidence sentences that were extracted from the papers.

In our series of experiments, we first evaluate the performance of a standard BERT-based classifier on the EICAT dataset, where full-text inputs must be split into multiple segments due to the limited context length. We then compare its performance to ModernBERT \cite{warner2024modernBERT}, a recent BERT variant capable of handling longer contexts, as well as to Llama-3.1 8B \cite{grattafiori2024llama3herdmodels}, a state-of-the-art local LLM.

All models face significant challenges due to the large input size, leading us to performing further experiments with training selector models to identify and prioritize the most relevant input sentences for training and evaluation. To ensure the general applicability of this approach, we test various sentence selection strategies, including leveraging human-provided evidence annotations, using LLM-generated selections, and using importance scores extracted from the classifiers.

Our findings indicate that many selection strategies improve classifier performance while simultaneously enhancing the efficiency of the decoder model. For scenarios where efficiency is less critical, we also observe that repeated randomization can improve the classification performance and even make a random selection of input sentences a viable strategy, thus leading to a simple-to-use way of boosting classification results.

Ultimately, this work presents a generalizable pipeline for accelerating inference and improving performance of scientific full-text classification.

The remainder of the paper is organized as follows: Section \ref{sec:2} reviews recent natural language processing approaches for the automated processing of scientific texts. Section \ref{sec:3} introduces the EICAT dataset, while Sections \ref{sec:4} and \ref{sec:5} describe our experiments and present the results. Section \ref{sec:6} provides a discussion of the findings, and Section \ref{sec:7} concludes with final remarks.

\section{Related Work}
\label{sec:2}

\subsection{Language Models for Scientific Literature}

The introduction of the transformer architecture \cite{attentionisallyouneed} revolutionized natural language processing, marking a new era in the field, with pretrained language models like BERT \cite{devlin-etal-2019-bert} significantly advancing performance across a wide range of tasks. This progress quickly extended to the scientific domain, leading to the development of domain-specific models such as SciBERT \cite{beltagy2019scibert}, which set new benchmarks on various scientific NLP tasks. SciBERT and similar models demonstrate clear advantages over general-purpose models \cite{biobert, song-etal-2023-matsci, rostam2024finetuninglargelanguagemodels}, and have therefore been applied to a variety of tasks within the scientific domain, including literature search and similarity assessment \cite{singh-etal-2023-scirepeval}, classification \cite{rostam2024finetuninglargelanguagemodels}, and summarization \cite{summarization}, with similar pretrained models having been trained for the general biomedical domain \cite{biobert, pubmedbert} as well as for the biodiversity domain \cite{abdelmageed2023biodivbert}.

More recently, the improved performance of autoregressive language models \cite{radford2019language} has driven a shift toward leveraging these models for a wide range of tasks. Openly available models, such as Llama-2 \cite{touvron2023llama2openfoundation}, alongside proprietary systems like ChatGPT, have established new state-of-the-art results in various scientific document processing tasks, including structured information extraction \cite{Rettenberger, dagdelen2024structured}, term extraction \cite{huang2024critical}, text classification, named entity recognition and and question answering \cite{choi2024accelerating}.

A range of benchmarks has been developed specifically for information extraction from scientific full texts, often accompanied by proposed models. These benchmarks target various tasks, including dataset mention detection \cite{dataset}, entity and relation extraction \cite{zhang-etal-2024-scier}, general information extraction \cite{jain-etal-2020-scirex}, and summarization \cite{deyoung-etal-2021-ms}.

\subsection{Language Models for Biodiversity Science}

In the specific domain of biodiversity science, transformer encoder architectures have been employed to tackle tasks such as hypothesis classification \cite{brinner-etal-2022-linking}, biodiversity analysis \cite{barcodebert}, named entity recognition and relation extraction \cite{abdelmageed2023biodivbert}, as well as hypothesis evidence localization \cite{weakClaims, brinner-zarriess-2024-rationalizing}. Additionally, autoregressive models have been applied to tasks such as literature review, question answering \cite{biodiversity_LLM}, and structured information extraction \cite{structuredExtractionEcology, infoextractBiodiv}, with further potential applications continuing to emerge \cite{osawa2023role}.

\section{The EICAT Dataset}
\label{sec:3}

We present a new dataset for training and evaluating models on the task of assessing the impact of invasive species on ecosystems based on scientific full texts. This dataset is grounded in the “Environmental Impact Classification for Alien Taxa” (EICAT, \citet{iucn2020guidelines}) standard, a classification standard developed by the International Union for Conservation of Nature (IUCN), which is used by researchers to compile standardized summaries of scientific literature addressing invasive species, along with assessments of the species' impacts as reported in these publications. The impacts are categorized into one of six possible classes: \textit{Minimal Concern}, \textit{Minor}, \textit{Moderate}, \textit{Major Risk}, \textit{Massive}, and \textit{Data Deficient}. Furthermore, researchers extract and include sentences from the full texts as evidence supporting their selected category. These impact assessments for various species are publicly available as Excel files at \href{https://www.iucngisd.org/gisd/}{https://www.iucngisd.org/gisd/}.

To construct our dataset, we acquired impact assessment files for as many species as possible. From these files, we extracted publication names and corresponding impact assessments for each species, covering around 800 publications. Using Llama-3 8B, we determined whether each citation represented a scientific paper (as opposed to books, PhD theses, government reports, etc.), since this study focuses exclusively on shorter scientific articles. We then used Crossref (\href{https://www.crossref.org/}{crossref.org}) as well as manual scraping to obtain as many full texts as possible.

Since the retrieved documents were in PDF format, we used Grobid \cite{GROBID} to extract the raw text from the publications. We excluded any documents for which text extraction was unsuccessful, resulting in a final dataset with 436 full texts addressing 120 species.

As a final processing step, we matched the evidence sentences from the impact assessments to sentences in the extracted text files. Discrepancies between the version of the paper we obtained and the one used for the assessments, as well as artifacts introduced during the PDF-to-text conversion, made exact matching infeasible in many cases. To address this, we implemented a fuzzy matching strategy, matching two sentences if they contain most of the same words in the same order. For matches slightly below the set similarity threshold, we used Llama-3 to determine whether the sentences were still a valid match. In total, we identified 2,247 evidence sentences, compared to 2,226 sentences in the original annotations. The higher count likely results from imperfect matching, as well as from PDF-to-text conversion artifacts, which sometimes split evidence sentences in the full text into two parts.

We created training, validation, and test splits comprising 82\%, 8\%, and 10\% of the species, respectively. To prevent inflated performance scores caused by the model learning the typical impact category assigned to a specific species across publications, all texts addressing the same species were assigned to the same split.

We publish the dataset containing publication names, impact labels and evidence sentences together with our code on \href{https://github.com/inas-argumentation/efficient_full_text_classification}{github.com/inas-argumentation/efficient\_full\_text\_classification}.

\begin{figure*}[h]
\fbox{
\begin{minipage}{0.99\textwidth}
\small{
This is a scientific paper about an invasive species: [SCIENTIFIC FULL TEXT]
\vspace{1em}

This is the end of the scientific text. Your task is to classify the impact that the invasive species [SPECIES NAME] has. Note that the text might contain information on other species. Possible classes are the following:

1. Minimal

A taxon is considered to have impacts of Minimal Concern when it causes negligible levels of have impacts on the recipient environment at some level, for example by altering species diversity or community similarity (e.g., biotic homogenisation), and for this reason there is no category equating to “no impact”. Only taxa for which changes in the individual performance of natives have been studied but not detected are assigned an MC category. Taxa that have been evaluated under the EICAT process but for which impacts have not been assessed in any study should not be classified in this category, but rather should be classified as Data Deficient.

2. Minor

A taxon is considered to have Minor impacts when it causes reductions in the performance of individuals in the native biota, but no declines in native population sizes, and has no impacts that would cause it to be classified in a higher impact category.

3. Moderate

A taxon is considered to have Moderate impacts when it causes declines in the population size of at least one native taxon, but has not been observed to lead to the local extinction of a native taxon.

4. Major

A taxon is considered to have Major impacts when it causes community changes through the local or sub-population extinction (or presumed extinction) of at least one native taxon, that would be naturally reversible if the alien taxon was no longer present. Its impacts do not lead to naturally irreversible local population, sub-population or global taxon extinctions.

5. Massive

A taxon is considered to have Massive impacts when it causes naturally irreversible community changes through local, sub-population or global extinction (or presumed extinction) of at least one native taxon.

6. Data Deficient

A taxon is categorised as Data Deficient when the best available evidence indicates that it has (or had) individuals existing in a wild state in a region beyond the boundary of its native geographic range, but either there is inadequate information to classify the taxon with respect to its impact, or insufficient time has elapsed since introduction for impacts to have become apparent. It is expected that all introduced taxa will have an impact at some level, because by definition an alien taxon in a new environment has a nonzero impact. However, listing a taxon as Data Deficient recognises that current information is insufficient to assess that level of impact.
\vspace{1em}

Return just the classification and end your answer, and provide one of the following labels as answer: "Minimal", "Minor", "Moderate", "Major", "Massive", "Data Deficient". Provide your answer by just using the following response format, and do not answer anything else in addition to that:

Summary: [One sentence summarizing the key information that you consider for the assessment]

Answer: [Your answer, that is one of the six labels]

END.}
\end{minipage}}
\caption{The Llama-3.1 8B prompt to classify scientific full texts.}
\label{fig:text3}
\vspace{-10px}
\end{figure*}

\section{Baseline Classification Experiments}
\label{sec:4}


\begin{table*}
\begin{center}
\begin{tabular*}{0.95\textwidth}{@{\extracolsep{\fill}}llcccc@{}}
\hline& & \multicolumn{2}{c}{\bf Deterministic Selection} & \multicolumn{2}{c}{\bf Randomized Selection} \\
\bf Model & \bf Sentence Selector & \bf Macro F1 & \bf Micro F1 & \bf Macro F1 & \bf Micro F1 \\ \hline
ModernBERT & Complete Input & 0.433 & 0.446 & 0.439 & 0.465 \\ \hline
PubMedBERT & Complete Input & 0.425 & 0.446 & - & - \\
PubMedBERT & Evidence & \bf 0.523 & \bf 0.538 & 0.503 & 0.508 \\
PubMedBERT & LLM & 0.457 & 0.460 & 0.494 & \underline{0.508} \\
PubMedBERT & Entropy & 0.453 & 0.460 & 0.475 & 0.494 \\
PubMedBERT & Importance & 0.442 & 0.460 & 0.479 & 0.479 \\
PubMedBERT & Random & 0.441 & 0.450 & \underline{0.496} & 0.499 \\ \hline
Llama 3.1 8B & Complete Input & 0.272 & 0.373 & - & - \\
Llama 3.1 8B & Evidence & 0.234 & 0.237 & 0.257 & 0.271 \\
Llama 3.1 8B & LLM & 0.228 & 0.271 & 0.230 & 0.305 \\
Llama 3.1 8B & Entropy & 0.322 & 0.373 & 0.358 & 0.441 \\
Llama 3.1 8B & Importance & 0.403 & 0.441 & 0.399 & 0.441 \\
Llama 3.1 8B & Random & 0.265 & 0.339 & 0.356 & 0.407 \\
\hline
\end{tabular*}
\end{center}
\caption{\label{score-table} Results on the EICAT dataset using PubMedBERT, ModernBERT and Llama-3.1 8B with either the full text input or one of the sentence selectors. Best scores are bold, second-best (from a different model) are underlined.}\vspace{-6px}
\end{table*}\vspace{-0px}

\subsection{Experimental Setup}

Our initial experiment focuses on establishing baseline performance for both BERT-based models and local instruction-tuned LLMs on our datasets. Specifically, we evaluate 1) PubMedBERT \cite{pubmedbert}, which demonstrated strong performance in previous studies on invasion biology \cite{brinner-etal-2022-linking} 2) ModernBERT \cite{warner2024modernBERT}, a recently introduced BERT variant that claims improved performance and efficiency while allowing for input lengths of up to 8192 tokens, and 3) Llama-3.1 8B, a state-of-the-art local LLM capable of handling up to 128K tokens, allowing for full-text processing.

Given that BERT-based models are limited to processing 512 tokens at a time, we split each full text into chunks of up to 512 tokens (with neighboring chunks overlapping for 50 tokens) and average the output logits across all chunks to produce a single score for the entire paper, serving as input to the cross-entropy loss. For ModernBERT, we instead used the whole full-text as input to the model, with tokens exceeding the 8192 token context window being truncated. We perform seven runs per model to obtain average results that mitigate variance in our reported scores.

For the LLM, we design a prompt that includes the full text along with a textual description of the impact categories, extracted from the IUCN EICAT guidelines \cite{iucn2020guidelines}. The model is prompted to first generate a sentence summarizing the impact (a step that significantly improves classification results) and then output a single impact category in a structured way (see Figure \ref{fig:text3}). We used greedy decoding to deterministically generate the most likely answer.

\subsection{Results}

The results of the classification experiment are presented in Table \ref{score-table} (Deterministic Selection, Sentence Selector: \textit{Complete Input}), where we report both macro F1 and micro F1 scores. Given the dataset's highly uneven label distribution, macro F1 can be strongly influenced by the misclassification of only a few samples, making micro F1 an important complementary metric. The results show that the trained BERT model achieves a macro F1 score of 0.425, thus significantly outperforming the LLM with a rather unsatisfactory macro F1 result of 0.272. We hypothesize that this could be cause by two key factors:

\begin{enumerate}
    \item Limited context in model prompt: While researchers use the same EICAT impact class descriptions as provided in the prompt for their assessments, they also rely on their domain knowledge and familiarity with existing literature to perform impact assessments as intended. A trained model learns this implicit consensus through exposure to a large, annotated dataset, whereas the LLM lacks this resource and depends solely on the textual descriptions of the classes, which are less informative.
    \item Challenges with input length: Full texts contain extensive information, not all of which will be relevant for the classification, thus making it hard to detect the relevant pieces of information to perform the classification.
\end{enumerate}

While the first issue could be addressed by training the LLM on the dataset, this is beyond the scope of this initial analysis. The second issue is supported by the fact that ModernBERT outperformed the standard BERT variant only marginally, even though it is able to reason about much more information at once, thus again indicating that the abundance of information in a full-text can pose significant challenges. To investigate this issue further, we conducted additional experiments that focus on selecting a subset of relevant sentences during preprocessing and using only those as input for the BERT or Llama models.

\section{Evidence Sentence Selection}
\label{sec:5}
\subsection{Experimental Setup}

We propose a two-step procedure to improve the performance of the models tested in the previous experiments. Our hypothesis is that both models face challenges due to the length of the full-text inputs. For the LLM, identifying the few critical pieces of information within a large block of text can be difficult. For PubMedBERT, the input is often split into more than 15 chunks, many of which might contain little to no relevant information, potentially disrupting the training process.

The proposed procedure involves training a sentence selector model (also based on PubMedBERT) on all sentences from the training set to distinguish important sentences from less relevant ones. Once trained, we use the selector to identify the 15 most relevant sentences for each document. Both models are then trained and evaluated using only these selected sentences, thus significantly reducing the input size while focusing on the most crucial information. For the Llama model, we used the same prompt as before, with the addition of mentioning at the beginning that sentences extracted from a paper are presented, and indicating left-out sentences in the input with "[...]", which lead to improved results.

We evaluate several strategies for training the sentence selector model:
\begin{enumerate}
    \item \textit{Evidence}: A model trained to recognize the human evidence annotations from the dataset.
    \item \textit{LLM}: We provide the EICAT guidelines as background and prompt Llama-3.1 8B to assess each individual sentence from a paper, classifying it as \textit{Not Useful}, \textit{Slightly Useful}, or \textit{Highly Useful}, resulting in a three-class classification task.
    \item \textit{Entropy}: We used three of the seven BERT classifiers trained in Section \ref{sec:4} to classify each sentence individually. A low entropy in the predicted distribution is a sign that the sentence is indicative of a specific class.
    \item \textit{Importance}: For each sentence in the dataset, we used three of the seven BERT classifiers trained in the earlier experiments to classify the corresponding full text, once with the sentence included and once with it being removed. We then evaluated the absolute change in output logits to assess the importance of the given sentence for the output.
\end{enumerate}

The evidence and LLM-based annotations naturally give rise to two- and three-class classification tasks for training the BERT sentence selection model. The entropy and importance scores, on the other hand, are continuous by nature, but since the absolute values of these scores are less relevant compared to the ranking among sentences within a text, we decided to discretize them into three categories: sentences falling within the bottom 50\% of scores within a text, those in the top 20\%, and the remaining 30\% in between, thus again constituting a three-class classification problem.

The sentence selectors all receive the species name and the three sentences before and after the sentence that they shall assess as context, with the relevant sentence being enclosed by \textit{[SEP]}-tokens. The resulting models can be used for ranking sentences within a document by predicting class probabilities for each sentence individually, and then using the expected value as continuous score.

\subsection{Results}

\subsubsection{Sentence Selector Agreement}

We begin by comparing the similarities between the predictions of the different sentence selector models (displayed in Table \ref{sentence-table}) to see if they focus on similar kinds of information. To quantify this, we use the normalized discounted cumulative gain (NDCG), which produces a score between 0 and 1, with higher values indicating greater agreement between the rankings of two models (i.e., highly ranked sentences by one model are also ranked highly by the other model or ground truth).

The rankings generated by the different trained sentence selector models are compared to the test-set ground truth rankings (i.e., the evidence annotations created by human annotators, or the assessments that were directly predicted by the LLM). Notably, the model trained on human evidence annotations achieved only a mediocre NDCG score with regards to alignment with the ground truth evidence annotations. This could be caused by inconsistencies in how evidence sentences were selected across different EICAT assessments, which might be caused by the involvement of many different researchers in their creation. Notably, it proved to be important to provide the surrounding sentences as well as the species name as context, since otherwise the NDCG score drops to just 0.487. The reason for this is, that annotated evidence sentences usually report on actual evidence collected within a study, so that the model needs to learn to exclude sentences that appear, for example, in a literature review section, which can be hard if that sentence is viewed in isolation. Further, a text might address several species, thus making the species for which the sentence shall be assessed a crucial piece of information.

In contrast to the evidence selector, the LLM demonstrated a high degree of internal consistency, achieving an impressive NDCG score of 0.911 between its own test set predictions and those from the corresponding BERT classifier.

Interestingly, while human and LLM rankings show some correlation, the two BERT-based methods for generating sentence rankings align only marginally better with the human or LLM annotations than a random selector. This raises concerns about the validity of these methods. However, their actual utility for the classification will be further evaluated in the following section.

\begin{table}[t]
\begin{center}
\begin{tabular}{lcc}
\hline \bf Train Data & \bf Evidence NDCG & \bf LLM NDCG \\ \hline
Evidence & 0.541 & 0.753 \\
LLM & 0.394 & 0.911 \\
Entropy & 0.362 & 0.691 \\
Importance & 0.344 & 0.674 \\
Random & 0.299 & 0.618 \\
\hline
\end{tabular}
\vspace{-6px}
\end{center}
\caption{NDCG scores denoting the match between the different sentence selection strategies and the ground truth sentences from the human evidence annotations or the LLM selections.}
\vspace{-6px}
\label{sentence-table}
\end{table}

\subsubsection{BERT Classification Results}

We evaluate the classification performance of BERT classifiers trained on the 15 most important sentences from each full text, as determined by the various sentence selectors. In most cases, only up to five sentences were chosen as evidence by the human annotators, but we chose the larger number of 15 to increase the likelihood of many relevant sentences being selected even if the selectors perform suboptimal, while still reducing the input size significantly. The results are presented in Table \ref{score-table} (Deterministic Selection).

For BERT classifiers, the evidence-based selector proves to be the most effective, significantly improving classification performance. A possible explanation is that it removes unnecessary and distracting information, most importantly because it can filter out sentences describing impacts caused by other species, thereby eliminating misleading information and implicitly creating a focus on the target species. In contrast, the BERT and ModernBERT models trained on the full input did not receive the species name, which was necessary to ensure they relied on textual evidence rather than simply associating species names with specific classifications, but leading to potentially incorrect predictions in the case of multiple species being addressed in a text. Since the BERT classifier was used as basis for training the importance and entropy selectors, these models likely did not learn to filter out sentences about other species as well. However, the LLM-based selector may have developed this ability, as the species name was included when generating the LLM assessments used for training. Nevertheless, it only marginally outperforms the entropy and importance selectors.

Overall, all sentence selection strategies improve classification performance, even when compared to ModernBERT, which should have access to the same (and even more) information. This holds even true for a random selection strategy, which selects 15 sentences before a training run starts and does not change this predetermined selection to mirror the deterministic selection by the other models. We see this as evidence that an overflow of information decreases classification performance, thus making our sentence selection strategy highly effective.

\subsubsection{Llama Classification Results}

The results for Llama reveal a different pattern compared to BERT. Despite their potentially beneficial property of filtering our non-relevant impacts, the evidence and LLM selectors do not improve classification performance. In this case, these properties will not be as significant, though, since the LLM does receive the name of the species it shall assess, so that it can filter out unnecessary information on its own.

To explain the decreased performance, we analyzed the distribution of class predictions and found that, for the evidence selector, the model's predictions significantly under-represent the lower-impact classes (\textit{Data Deficient}, \textit{Minimal Concern} and \textit{Minor}). We attribute this to the model receiving condensed information on the impact of the specific invasive species, thus pushing it to a higher impact category that it did not see as justified when assessing the full text.

For the LLM selector, we see a similar distribution, with a few more samples being classified as \textit{Minor}, but even less being classified as \textit{Data Deficient}, which could be caused by the LLM not managing to exclude sentences from, for example, the literature review section, thus making every paper contain some information on potential impacts.

The BERT model, in contrast, is not susceptible to these factors hindering the LLM, since it is additionally trained on these specific inputs and thus learns to draw the right conclusion from them.

Interestingly, sentences identified as important by BERT (i.e., \textit{Entropy} and \textit{Importance}) lead to substantially better results than the other strategies. We see this as a sign that the models indeed learned to identify the sentences that should actually contribute to the classification (as learned by the original BERT model), thus mitigating especially the issues pointed out for the LLM-selector.

\subsubsection{Randomization}

In both experiments, random sentence selection yielded reasonable results, even outperforming using the complete input or other selection methods. This is especially notable for the BERT models, as we fixed the 15 randomly selected sentences for each sample within a training run, thus significantly restricting access to useful information. A similar limitation applies to the other selection strategies, which also reduce the total number of sentences encountered during training.

To explore this further, we conducted additional experiments where a new random input is created each time a text is accessed during training. For the targeted selectors, this random sampling is restricted to the top 30 sentences (with higher-ranking sentences being sampled more often), ensuring that most sentences deemed unimportant were excluded. During evaluation, we generated 10 different input samples per text and determined the final prediction through majority voting.

The results for the randomized classifiers are shown in Table \ref{score-table} (Randomized Selection). With the exception of the evidence selector, we observe consistent performance improvements across all BERT models. This suggests that, unless the selection is guided by a well-informed approach based on human annotations, exposing the model to a greater variety of sentences during training and making predictions based on diverse inputs is beneficial. Notably, this even makes the random selector a viable competitor to the evidence selector, demonstrating that for large-text classification tasks, this simple strategy can be an effective choice. We hypothesize that targeted selectors focus on specific types of information (such as empirical observations for the evidence selector), which leads to only very narrow relationships being learned during training. In contrast, the random selector’s lack of bias increases the variance of inputs and forces the model to generalize more effectively through being trained on a more difficult task, thus enabling it to learn the broader relationships required for accurate classification.

For the Llama model, randomization improves results across most selection strategies, with the random selector again becoming a viable alternative to targeted selection. We hypothesize that specific sentences might throw off the LLM's prediction for a specific input, and sampling many different inputs could instead lead to a classification that is based on the general information provided by a vast number of sentences in the text.

\subsubsection{Efficiency Analysis}

In the introduction, we emphasized the importance of efficiency for broader adoption of local models outside the machine learning community. Alongside performance improvements, we observed substantial speed gains for the Llama model due to reduced input lengths. For instance, a full test set evaluation with full-text input on an RTX 3090 takes 116 seconds, but this drops to 65 seconds with importance-based sentence selection - including the time for sentence relevance prediction. Notably, this strategy also improves classification performance, breaking the typical trade-off between efficiency and accuracy. Further increased performance using sampling then leads to vastly increased times that are more than three times longer than using the full-text input.

Smaller models like BERT remain far more efficient, requiring just 6.6 seconds for a test set evaluation with evidence sentence selection. The sampling strategy increases this to 9.8 seconds, thus offering performance gains for most selection strategies in trade for higher computational cost.

\section{Discussion}
\label{sec:6}

In our evaluation, we identified significant challenges when using instruction-tuned LLMs for scientific text processing. On the one hand, extracting a different set of sentences, even if they should contain the necessary information for performing the classification, can easily change classification results and even push the model towards incorrect conclusions. Additionally, a detailed natural language description of our task was insufficient for the Llama model to achieve results comparable to a 70 times smaller BERT classifier, and the explicit selection of highly relevant sentences through the evidence selector did not yield improvements. We interpret this as a sign that sample-level labels, as used by BERT, provide substantially more information than both evidence annotations and natural language descriptions.

The superior informational content of sample-level labels compared to evidence annotations is plausible considering that a single evidence annotation only conveys information about a single sentence, while a sample-level label provides information about every sentence in the whole text. On the other hand, the superior performance of sample-level labels over natural language descriptions is especially significant given the recent trend toward prompting-based approaches rather than extensive labeling efforts. In-context learning \cite{dong2024surveyincontextlearning} offers a potential bridge between these approaches, enabling the delivery of rich sample-level information to LLMs without training, typically complementing natural language descriptions within the prompting framework. This combination can thus potentially overcome the challenge of precisely specifying a given task by returning to the classical way of demonstrating desired behavior. However, while this approach has shown success, it becomes impractical for tasks involving lengthy inputs, such as scientific full-text classification. For such cases, fine-tuning local large language models could present a viable solution, which can be explored in future work.

\section{Conclusion}
\label{sec:7}
We introduced a novel dataset for scientific full-text classification and conducted extensive experiments using smaller encoder and larger decoder architectures. Our results demonstrated that various strategies for reducing input size can simultaneously enhance efficiency and performance, offering a generalizable pipeline adaptable to other tasks. However, as classification scores remain suboptimal, future research could investigate the potential of fine-tuning local LLMs, leveraging recently emerging LLMs with advanced reasoning capabilities \cite{deepseekai2025deepseekr1incentivizingreasoningcapability}, or testing the performance of larger proprietary models.


\bibliographystyle{acl_natbib}
\bibliography{main}

\begin{thebibliography}{37}
\expandafter\ifx\csname natexlab\endcsname\relax\def\natexlab#1{#1}\fi

\bibitem[{GRO(2008--2024)}]{GROBID}
 2008--2024.
\newblock \href {http://arxiv.org/abs/1:dir:dab86b296e3c3216e2241968f0d63b68e8209d3c} {Grobid}.
\newblock \url{https://github.com/kermitt2/grobid}.

\bibitem[{Abdelmageed et~al.(2023)Abdelmageed, L{\"o}ffler, and K{\"o}nig-Ries}]{abdelmageed2023biodivbert}
Nora Abdelmageed, Felicitas L{\"o}ffler, and Birgitta K{\"o}nig-Ries. 2023.
\newblock Biodivbert: a pre-trained language model for the biodiversity domain.
\newblock In \emph{SWAT4HCLS}, pages 62--71.

\bibitem[{Arias et~al.(2023)Arias, Sadjadi, Safari, Gong, Wang, Lowe, Haurum, Zarubiieva, Steinke, Kari, Chang, and Taylor}]{barcodebert}
Pablo~Millan Arias, Niousha Sadjadi, Monireh Safari, ZeMing Gong, Austin~T. Wang, Scott~C. Lowe, Joakim~Bruslund Haurum, Iuliia Zarubiieva, Dirk Steinke, Lila Kari, Angel~X. Chang, and Graham~W. Taylor. 2023.
\newblock \href {http://arxiv.org/abs/2311.02401} {Barcodebert: Transformers for biodiversity analysis}.

\bibitem[{Beltagy et~al.(2019)Beltagy, Lo, and Cohan}]{beltagy2019scibert}
Iz~Beltagy, Kyle Lo, and Arman Cohan. 2019.
\newblock \href {https://doi.org/10.18653/v1/D19-1371} {{S}ci{BERT}: A pretrained language model for scientific text}.
\newblock In \emph{Proceedings of the 2019 Conference on Empirical Methods in Natural Language Processing and the 9th International Joint Conference on Natural Language Processing (EMNLP-IJCNLP)}, pages 3615--3620, Hong Kong, China. Association for Computational Linguistics.

\bibitem[{Bornmann et~al.(2021)Bornmann, Haunschild, and Mutz}]{bornmann_growth_2021}
Lutz Bornmann, Robin Haunschild, and Rüdiger Mutz. 2021.
\newblock \href {https://doi.org/10.1057/s41599-021-00903-w} {Growth rates of modern science: a latent piecewise growth curve approach to model publication numbers from established and new literature databases}.
\newblock 8(1):1--15.
\newblock Publisher: Palgrave.

\bibitem[{Brinner et~al.(2022)Brinner, Heger, and Zarriess}]{brinner-etal-2022-linking}
Marc Brinner, Tina Heger, and Sina Zarriess. 2022.
\newblock \href {https://doi.org/10.18653/v1/2022.wiesp-1.5} {Linking a hypothesis network from the domain of invasion biology to a corpus of scientific abstracts: The {INAS} dataset}.
\newblock In \emph{Proceedings of the first Workshop on Information Extraction from Scientific Publications}, pages 32--42, Online. Association for Computational Linguistics.

\bibitem[{Brinner et~al.(2024)Brinner, Zarrie{\ss}, and Heger}]{weakClaims}
Marc Brinner, Sina Zarrie{\ss}, and Tina Heger. 2024.
\newblock Weakly supervised claim localization in scientific abstracts.
\newblock In \emph{Robust Argumentation Machines}, pages 20--38, Cham. Springer Nature Switzerland.

\bibitem[{Brinner and Zarrie{\ss}(2024)}]{brinner-zarriess-2024-rationalizing}
Marc~Felix Brinner and Sina Zarrie{\ss}. 2024.
\newblock \href {https://doi.org/10.18653/v1/2024.emnlp-main.664} {Rationalizing transformer predictions via end-to-end differentiable self-training}.
\newblock In \emph{Proceedings of the 2024 Conference on Empirical Methods in Natural Language Processing}, pages 11894--11907, Miami, Florida, USA. Association for Computational Linguistics.

\bibitem[{Castro et~al.(2024)Castro, Pinto, Reino, Pipek, and Capinha}]{structuredExtractionEcology}
Andry Castro, João Pinto, Luís Reino, Pavel Pipek, and César Capinha. 2024.
\newblock \href {https://doi.org/https://doi.org/10.1016/j.ecoinf.2024.102742} {Large language models overcome the challenges of unstructured text data in ecology}.
\newblock \emph{Ecological Informatics}, 82:102742.

\bibitem[{Choi and Lee(2024)}]{choi2024accelerating}
Jaewoong Choi and Byungju Lee. 2024.
\newblock Accelerating materials language processing with large language models.
\newblock \emph{Communications Materials}, 5(1):13.

\bibitem[{Dagdelen et~al.(2024)Dagdelen, Dunn, Lee, Walker, Rosen, Ceder, Persson, and Jain}]{dagdelen2024structured}
John Dagdelen, Alexander Dunn, Sanghoon Lee, Nicholas Walker, Andrew~S Rosen, Gerbrand Ceder, Kristin~A Persson, and Anubhav Jain. 2024.
\newblock Structured information extraction from scientific text with large language models.
\newblock \emph{Nature Communications}, 15(1):1418.

\bibitem[{DeepSeek-AI et~al.(2024)}]{deepseekai2024deepseekv3technicalreport}
DeepSeek-AI et~al. 2024.
\newblock \href {http://arxiv.org/abs/2412.19437} {Deepseek-v3 technical report}.

\bibitem[{DeepSeek-AI et~al.(2025)}]{deepseekai2025deepseekr1incentivizingreasoningcapability}
DeepSeek-AI et~al. 2025.
\newblock \href {http://arxiv.org/abs/2501.12948} {Deepseek-r1: Incentivizing reasoning capability in llms via reinforcement learning}.

\bibitem[{Devlin et~al.(2019)Devlin, Chang, Lee, and Toutanova}]{devlin-etal-2019-bert}
Jacob Devlin, Ming-Wei Chang, Kenton Lee, and Kristina Toutanova. 2019.
\newblock \href {https://doi.org/10.18653/v1/N19-1423} {{BERT}: Pre-training of deep bidirectional transformers for language understanding}.
\newblock In \emph{Proceedings of the 2019 Conference of the North {A}merican Chapter of the Association for Computational Linguistics: Human Language Technologies, Volume 1 (Long and Short Papers)}, pages 4171--4186, Minneapolis, Minnesota. Association for Computational Linguistics.

\bibitem[{DeYoung et~al.(2021)DeYoung, Beltagy, van Zuylen, Kuehl, and Wang}]{deyoung-etal-2021-ms}
Jay DeYoung, Iz~Beltagy, Madeleine van Zuylen, Bailey Kuehl, and Lucy~Lu Wang. 2021.
\newblock \href {https://doi.org/10.18653/v1/2021.emnlp-main.594} {{MS}\^{}2: Multi-document summarization of medical studies}.
\newblock In \emph{Proceedings of the 2021 Conference on Empirical Methods in Natural Language Processing}, pages 7494--7513, Online and Punta Cana, Dominican Republic. Association for Computational Linguistics.

\bibitem[{Dong et~al.(2024)Dong, Li, Dai, Zheng, Ma, Li, Xia, Xu, Wu, Liu, Chang, Sun, Li, and Sui}]{dong2024surveyincontextlearning}
Qingxiu Dong, Lei Li, Damai Dai, Ce~Zheng, Jingyuan Ma, Rui Li, Heming Xia, Jingjing Xu, Zhiyong Wu, Tianyu Liu, Baobao Chang, Xu~Sun, Lei Li, and Zhifang Sui. 2024.
\newblock \href {http://arxiv.org/abs/2301.00234} {A survey on in-context learning}.

\bibitem[{Grattafiori et~al.(2024)}]{grattafiori2024llama3herdmodels}
Aaron Grattafiori et~al. 2024.
\newblock \href {http://arxiv.org/abs/2407.21783} {The llama 3 herd of models}.

\bibitem[{Gu et~al.(2021)Gu, Tinn, Cheng, Lucas, Usuyama, Liu, Naumann, Gao, and Poon}]{pubmedbert}
Yu~Gu, Robert Tinn, Hao Cheng, Michael Lucas, Naoto Usuyama, Xiaodong Liu, Tristan Naumann, Jianfeng Gao, and Hoifung Poon. 2021.
\newblock \href {https://doi.org/10.1145/3458754} {Domain-specific language model pretraining for biomedical natural language processing}.
\newblock \emph{ACM Trans. Comput. Healthcare}, 3(1).

\bibitem[{Huang et~al.(2024)Huang, Yang, Rong, Nezafati, Treager, Chi, Wang, Cheng, Guo, Klesse et~al.}]{huang2024critical}
Jingwei Huang, Donghan~M Yang, Ruichen Rong, Kuroush Nezafati, Colin Treager, Zhikai Chi, Shidan Wang, Xian Cheng, Yujia Guo, Laura~J Klesse, et~al. 2024.
\newblock A critical assessment of using chatgpt for extracting structured data from clinical notes.
\newblock \emph{npj Digital Medicine}, 7(1):106.

\bibitem[{IUCN(2020)}]{iucn2020guidelines}
IUCN. 2020.
\newblock Guidelines for using the iucn environmental impact classification for alien taxa (eicat) categories and criteria. version 1.1.

\bibitem[{Jain et~al.(2020)Jain, van Zuylen, Hajishirzi, and Beltagy}]{jain-etal-2020-scirex}
Sarthak Jain, Madeleine van Zuylen, Hannaneh Hajishirzi, and Iz~Beltagy. 2020.
\newblock \href {https://doi.org/10.18653/v1/2020.acl-main.670} {{S}ci{REX}: {A} challenge dataset for document-level information extraction}.
\newblock In \emph{Proceedings of the 58th Annual Meeting of the Association for Computational Linguistics}, pages 7506--7516, Online. Association for Computational Linguistics.

\bibitem[{Jiqi~Gu(2024)}]{biodiversity_LLM}
Jiangshan~Lai Jiqi~Gu, Jianping~Chen. 2024.
\newblock \href {https://doi.org/10.17520/biods.2024258} {Application of large language models in biodiversity research}.
\newblock \emph{Biodiversity Science}, 32(9):24258.

\bibitem[{Kommineni et~al.(2024)Kommineni, Ahmed, Koenig-Ries, and Samuel}]{infoextractBiodiv}
Vamsi~Krishna Kommineni, Waqas Ahmed, Birgitta Koenig-Ries, and Sheeba Samuel. 2024.
\newblock \href {https://doi.org/10.3897/biss.8.136735} {Automating information retrieval from biodiversity literature using large language models: A case study}.
\newblock \emph{Biodiversity Information Science and Standards}, 8:e136735.

\bibitem[{Lee et~al.(2019)Lee, Yoon, Kim, Kim, Kim, So, and Kang}]{biobert}
Jinhyuk Lee, Wonjin Yoon, Sungdong Kim, Donghyeon Kim, Sunkyu Kim, Chan~Ho So, and Jaewoo Kang. 2019.
\newblock \href {https://doi.org/10.1093/bioinformatics/btz682} {Biobert: a pre-trained biomedical language representation model for biomedical text mining}.
\newblock \emph{Bioinformatics}, 36(4):1234--1240.

\bibitem[{Osawa et~al.(2023)Osawa, Tsutsumida et~al.}]{osawa2023role}
T~Osawa, N~Tsutsumida, et~al. 2023.
\newblock The role of large language models in ecology and biodiversity conservation: Opportunities and challenges.

\bibitem[{Pan et~al.(2023)Pan, Zhang, Dragut, Caragea, and Latecki}]{dataset}
Huitong Pan, Qi~Zhang, Eduard Dragut, Cornelia Caragea, and Longin~Jan Latecki. 2023.
\newblock \href {https://doi.org/10.1162/tacl_a_00592} {Dmdd: A large-scale dataset for dataset mentions detection}.
\newblock \emph{Transactions of the Association for Computational Linguistics}, 11:1132--1146.

\bibitem[{Radford et~al.(2019)Radford, Wu, Child, Luan, Amodei, Sutskever et~al.}]{radford2019language}
Alec Radford, Jeffrey Wu, Rewon Child, David Luan, Dario Amodei, Ilya Sutskever, et~al. 2019.
\newblock Language models are unsupervised multitask learners.
\newblock \emph{OpenAI blog}, 1(8):9.

\bibitem[{Rettenberger et~al.(2024)Rettenberger, Münker, Schutera, Niemeyer, Rabe, and Reischl}]{Rettenberger}
Luca Rettenberger, Marc~F. Münker, Mark Schutera, Christof~M. Niemeyer, Kersten~S. Rabe, and Markus Reischl. 2024.
\newblock \href {https://doi.org/doi:10.1515/cdbme-2024-2129} {Using large language models for extracting structured information from scientific texts}.
\newblock \emph{Current Directions in Biomedical Engineering}, 10(4):526--529.

\bibitem[{Rostam and Kertész(2024)}]{rostam2024finetuninglargelanguagemodels}
Zhyar Rzgar~K Rostam and Gábor Kertész. 2024.
\newblock \href {http://arxiv.org/abs/2412.00098} {Fine-tuning large language models for scientific text classification: A comparative study}.

\bibitem[{Sefid and Giles(2022)}]{summarization}
Athar Sefid and C.~Lee Giles. 2022.
\newblock Scibertsum: Extractive summarization for scientific documents.
\newblock In \emph{Document Analysis Systems}, pages 688--701, Cham. Springer International Publishing.

\bibitem[{Singh et~al.(2023)Singh, D{'}Arcy, Cohan, Downey, and Feldman}]{singh-etal-2023-scirepeval}
Amanpreet Singh, Mike D{'}Arcy, Arman Cohan, Doug Downey, and Sergey Feldman. 2023.
\newblock \href {https://doi.org/10.18653/v1/2023.emnlp-main.338} {{S}ci{R}ep{E}val: A multi-format benchmark for scientific document representations}.
\newblock In \emph{Proceedings of the 2023 Conference on Empirical Methods in Natural Language Processing}, pages 5548--5566, Singapore. Association for Computational Linguistics.

\bibitem[{Singha~Roy and Mercer(2024)}]{singha-roy-mercer-2024-enhancing}
Sudipta Singha~Roy and Robert~E. Mercer. 2024.
\newblock Enhancing scientific document summarization with research community perspective and background knowledge.
\newblock In \emph{Proceedings of the 2024 Joint International Conference on Computational Linguistics, Language Resources and Evaluation (LREC-COLING 2024)}, pages 6048--6058, Torino, Italia. ELRA and ICCL.

\bibitem[{Song et~al.(2023)Song, Miret, and Liu}]{song-etal-2023-matsci}
Yu~Song, Santiago Miret, and Bang Liu. 2023.
\newblock \href {https://doi.org/10.18653/v1/2023.acl-long.201} {{M}at{S}ci-{NLP}: Evaluating scientific language models on materials science language tasks using text-to-schema modeling}.
\newblock In \emph{Proceedings of the 61st Annual Meeting of the Association for Computational Linguistics (Volume 1: Long Papers)}, pages 3621--3639, Toronto, Canada. Association for Computational Linguistics.

\bibitem[{Touvron et~al.(2023)}]{touvron2023llama2openfoundation}
Hugo Touvron et~al. 2023.
\newblock \href {http://arxiv.org/abs/2307.09288} {Llama 2: Open foundation and fine-tuned chat models}.

\bibitem[{Vaswani et~al.(2017)Vaswani, Shazeer, Parmar, Uszkoreit, Jones, Gomez, Kaiser, and Polosukhin}]{attentionisallyouneed}
Ashish Vaswani, Noam Shazeer, Niki Parmar, Jakob Uszkoreit, Llion Jones, Aidan~N Gomez, \L~ukasz Kaiser, and Illia Polosukhin. 2017.
\newblock Attention is all you need.
\newblock In \emph{Advances in Neural Information Processing Systems}, volume~30. Curran Associates, Inc.

\bibitem[{Warner et~al.(2024)Warner, Chaffin, Clavié, Weller, Hallström, Taghadouini, Gallagher, Biswas, Ladhak, Aarsen, Cooper, Adams, Howard, and Poli}]{warner2024modernBERT}
Benjamin Warner, Antoine Chaffin, Benjamin Clavié, Orion Weller, Oskar Hallström, Said Taghadouini, Alexis Gallagher, Raja Biswas, Faisal Ladhak, Tom Aarsen, Nathan Cooper, Griffin Adams, Jeremy Howard, and Iacopo Poli. 2024.
\newblock \href {http://arxiv.org/abs/2412.13663} {Smarter, better, faster, longer: A modern bidirectional encoder for fast, memory efficient, and long context finetuning and inference}.

\bibitem[{Zhang et~al.(2024)Zhang, Chen, Pan, Caragea, Latecki, and Dragut}]{zhang-etal-2024-scier}
Qi~Zhang, Zhijia Chen, Huitong Pan, Cornelia Caragea, Longin~Jan Latecki, and Eduard Dragut. 2024.
\newblock \href {https://doi.org/10.18653/v1/2024.emnlp-main.726} {{S}ci{ER}: An entity and relation extraction dataset for datasets, methods, and tasks in scientific documents}.
\newblock In \emph{Proceedings of the 2024 Conference on Empirical Methods in Natural Language Processing}, pages 13083--13100, Miami, Florida, USA. Association for Computational Linguistics.

\end{thebibliography}

\end{document}